\documentclass[5p]{elsarticle}
\makeatletter
\def\ps@pprintTitle{%
 \let\@oddhead\@empty
 \let\@evenhead\@empty
 \def\@oddfoot{\centerline{\thepage}}%
 \let\@evenfoot\@oddfoot}
\makeatother

\usepackage{lineno,hyperref}
\usepackage{amsmath,amssymb,amsfonts} 
\usepackage{array,booktabs} 
\usepackage[utf8]{inputenc}

\journal{Biosystems Engineering}

\begin{document}

\begin{frontmatter}

\title{An Adaptive Approach for Automated Grapevine Phenotyping using VGG-based Convolutional Neural Networks}

\author[bonn]{Jonatan Grimm}
\author[siebeldingen]{Katja Herzog}
\author[siebeldingen]{Florian Rist}
\author[siebeldingen]{Anna Kicherer}
\author[siebeldingen]{Reinhard T\"opfer}
\author[bonn]{Volker Steinhage\corref{correspondingauthor}}

\cortext[correspondingauthor]{Corresponding author, Tel.: +49 (0)228 734538}
\ead{steinhage@cs.uni-bonn.de (Volker Steinhage)}

\address[bonn]{Department of Computer Science IV, University of Bonn, Endenicher Allee 19A, D-53115 Bonn, Germany}
\address[siebeldingen]{Institute for Grapevine Breeding Geilweilerhof, Julius K\"uhn-Institut, Federal Research Centre for Cultivated Plants, Siebeldingen, Germany}

\begin{abstract}
In (grapevine) breeding programs and research, periodic phenotyping and multi-year monitoring of different grapevine traits, like growth or yield, is needed especially in the field. This demand imply objective, precise and automated methods using sensors and adaptive software. This work presents a proof-of-concept analyzing RGB images of different growth stages of grapevines with the aim to detect and quantify promising plant organs which are related to yield.
\\ \\
The input images are segmented by a Fully Convolutional Neural Network (FCN) into object and background pixels. The objects are plant organs like young shoots, pedicels, flower buds or grapes, which are principally suitable for yield estimation. In the ground truth of the training images, each object is separately annotated as a connected segment of object pixels, which enables end-to-end learning of the object features. Based on the CNN-based segmentation, the number of objects is determined by detecting and counting connected components of object pixels using region labeling.
\\ \\
In an evaluation on six different data sets, the system achieves an IoU of up to 87.3\% for the segmentation and an F1 score of up to 88.6\% for the object detection. 
\end{abstract}


\end{frontmatter}



\section{Introduction}
\label{sec:intro}
\begin{figure*}[bt]
  \centering
  \includegraphics[width=1.0\textwidth]{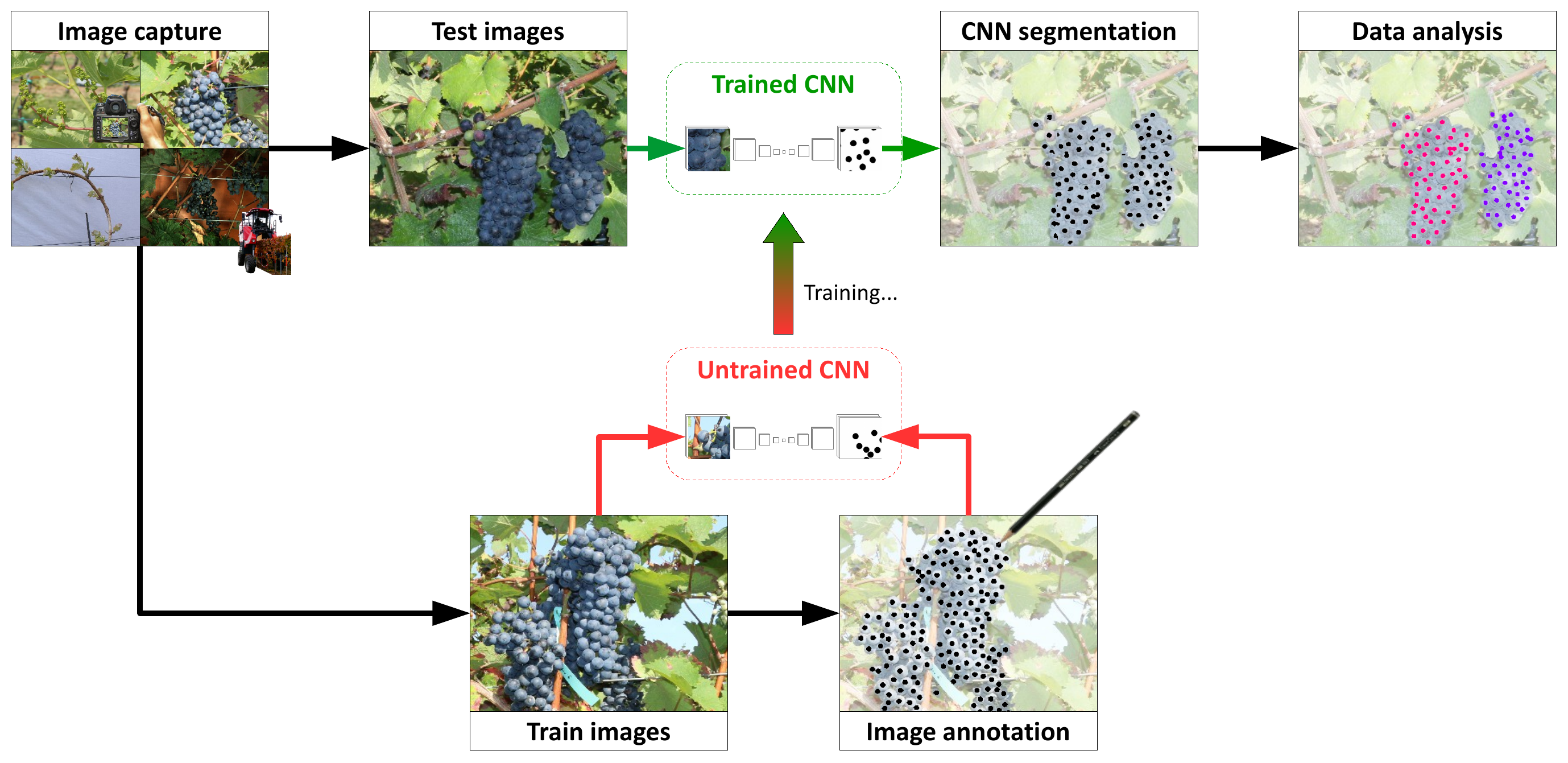}
  \caption{The architecture of the system developed for this work. }
  \label{fig:graphical_abstract}
\end{figure*}
Periodic and multi-scale field phenotyping of seedlings, mapping populations or genetic repositories are one of the most challenging demands of today’s grapevine breeding programs and research. Challenging because of large field sizes with several hundreds of plants that need to be phenotyped by laborious, manual and subjective classification methods (BBCH \cite{lorenz1995growth} or OIV \cite{oiv2009DescrList}) within a limited time schedule.\\ \\
In the last years, several tools using imaging sensors, field phenotyping platforms and automated image interpretation approaches were published showing its promising application for an objective, precise and partly high-throughput field phenotyping in order to overcome the existing phenotyping bottleneck (an overview is given by \citet{seng2018computer}, \citet{kicherer2017phenoliner}, \citet{kicherer2015automated}, \citet{rose2016towards}, \citet{herrero2015vineyard}). However, the implementation of such digital methods in breeding, research and wine industry requires robust, fast and adaptive tools operating reliably independent from the types or quality of acquired images from variable plant phenotypes and different traits of interest.\\ \\
There are already some existing approaches for automated yield estimation of vineyards. Many of them are destructive in nature and require plant organ collection before harvest followed by laborious measurements. A detailed comparison of such methods is given by \citet{de2015comparison}. Some approaches work on images taken in the field, so they are not destructive. Most of them work on images of single inflorescences with artificial backgrounds, like \citet{diago2014assessment} and \citet{millan2017image}.\\
The detection of single flower buds or unripe grapes in field images with a natural background is challenging, since the flower buds and unripe grapes are green as well as the leaves in the background. The colors can be affected by the sun light making it difficult to detect the objects by the color. Some approaches use flashlight images taken at night to avoid light artifacts. \citet{grossetete2012early} work on flashlight images taken with smart phones and make use of light reflections for berry detection. Such a reflection is typically a white circle in the center of the grapes and can be detected as a unique mountain peak in a Gaussian correlation map. \citet{nuske2014automated} work on flash light images taken with special vehicles which illuminate the grapes with big flashlights and take images with a camera. First, their approach searches for interest points which are candidates to be a grape. Interest points are detected by modeling the flash light reflections, but also by using the radial symmetry transform. Then each interest point is classified as berry or non-berry using a randomized KD forest. All of these applications require special illumination conditions, e.g. image acquisition at night or the application of very bright flashlight. Further, taking images of single inflorescences with an artificial background is very time-consuming.\\
\citet{rudolph2018efficient} work on unprepared field images of grapevines and generate a semantic segmentation using CNNs in order to detect inflorescences as regions of interest. In the detected regions of interest, the circular Hough transform (CHT) is applied to detect flower buds. A big advantage of this is that it works on unprepared images taken by consumer cameras during the day.\\
The present work is based on the work of \citet{rudolph2018efficient} and also uses a CNN-based approach, which is suitable for unprepared images. 
The main contribution of this work is the development of an approach for detection and localization of arbitrary plant organs in different growth stages. For this purpose, the CNN is not trained for ROI detection (like in \citet{rudolph2018efficient}), but directly in an end-to-end learning for the detection of the target objects, i.e., shoots, pedicels, flower buds, berries, etc. That means to use the CNN to generate an instance-level semantic segmentation. Thus, the need for a second processing step for object detection, like the CHT, is omitted. Only a region labeling is needed to detect and localize connected components, each representing an object (like a single grape berry). Hence there are no model assumptions about the type of target objects encoded in the segmentation process. Therefore, this segmentation process can be applied to the detection of arbitrary plant organs. Only new training of the CNN is neccessary to direct the CNN what to look for.\\
Specially for this task, it is very useful that the CNN architecture used in this work can be trained end-to-end. In general, end-to-end learning has the advantage, that the model can be directly learned as a global optimization, since no sub-networks must be pre-trained to produce the output. This enables a better generalization of the model and allows a better parallelization. For the task of object detection and localization, end-to-end learning allows that the CNN directly learns the features of the single objects, since they are annotated separately. This helps to achieve the objective, that the annotated images are the only model knowledge about the objects, which makes the approach suitable for detecting plant organs of arbitrary growth stages.\\ \\
%
Figure \ref{fig:graphical_abstract} shows the work flow of our system. As described in section \ref{sec:data_material}, the images can be captured unprepared with common consumer cameras, but the system also works on prepared artificial images. To be precise, a separate system is trained for each image data set, but these systems have the same architecture and only differ in their CNN model and in data-specific parameters like the size of the objects to be detected. The images are divided into training images and test images. The training images are used to train the CNN and have to be annotated by a human, since the CNN needs the ground truth of the images to learn its task (e.g. detecting grapes). After the CNN is trained, the test images can be segmented by the CNN. Based on this segmentation, the objects can be detected and localized by detecting connected components of object pixels using region labeling. This allows further data analysis like counting the objects, analyzing clusters or computing the length of the objects. This information can be used for yield estimation or other tasks in viticulture. \\ \\
This work introduces an approach for automated detection, localization, count and analysis (clustering and length computation) of objects like grapes, inflorescences and single flower buds, pedicels or young shoots, which are suitable to estimate the yield. The main idea of this approach is to use a Convolutional Neural Network (CNN) to generate a semantic segmentation of the images. This means that each pixel is classified as an object pixel or as a background pixel. Based on this segmentation, the objects can be localized and counted by searching for connected components of object pixels using region labeling. The semantic segmentation, as well as the object detection and localization, were tested and evaluated on six image data sets covering three different growth stages of grapevines.\\ 


\section{Material and Methods}

\subsection{Plant Material, Image Capture and Image Annotation}

\subsubsection{Plant Material and Image Capture}
\label{sec:data_material}

\begin{figure}[bth]

  \centering
  \includegraphics[width=0.48\textwidth]{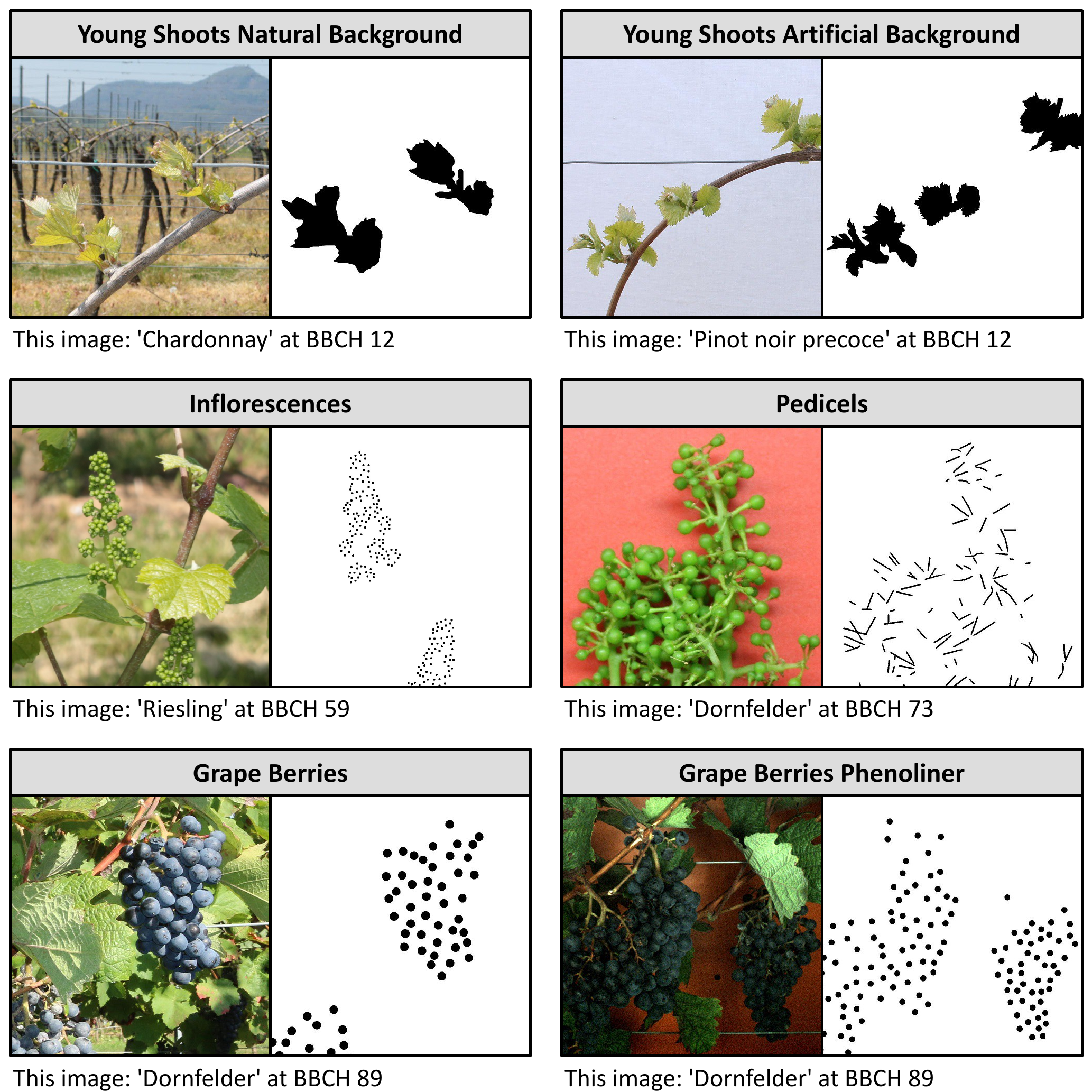}
  \caption{The six independent data sets of grapevine images used in this work. The left images show an excerpt of a training image, the right images show the corresponding ground truth which should be learned by the CNN. }
  \label{fig:data_material}
\end{figure}

The images used for the experiments in this work are from six data sets of grapevine images covering three different BBCH stages of grapevine development, namely young shoots at BBCH 12, inflorescences at BBCH 59 and ripe grapes at BBCH 89. The images were captured in 2017 at the experimental vineyards of Geilweilerhof located at Siebeldingen, Germany (Lat 49${}^\circ$13'2.892'', Lon 8${}^\circ$2'48.408''). Inter-row distance was 2 m, and grapevine spacing was 1 m. Six data sets of grapevine images were used in the present study: \\
Young shoots at BBCH 12: One image per grapevine and 10 grapevines per cultivar were captured (Aligote, Bacchus, Cabernet Sauvignon, Chardonnay, Grenache, Felicia, Fruehburgunder, Merlot, Müller-Thurgau, Pinot blanc, Pinot noir, Pinot noir precoce, Primitivo, Regent, Reberger, Riesling, Sauvignon Blanc, Silvaner, Syrah, Tempra, Villaris).\\
Inflorescences: Three images per grapevine and 3 grapevines per cultivar were captured (Chardonnay, Riesling).
Pedicels: Two images per inflorescence and 20 Inflorescences per cultivar were taken in the lab under standardized conditions (i.e. 40 images;  Riesling, Chardonnay, Dornfelder, Regent)\\
Ripe grapes at BBCH 89: One image per grapevine (21 plants per cultivar) was captured directly in the field by using consumer camera (SLR) or the automated phenotyping platform Phenoliner (\citet{kicherer2017phenoliner}).\\
Figure \ref{fig:data_material} shows representative image excerpts and their corresponding ground truth for each data set.
\\ \\
All data sets are divided into a mostly equal number of training and evaluation images. The remaining images are test images, which are not evaluated. The number of needed training images depends on the complexity of the objects to learn and on the average number of objects per image. Table \ref{table-number-objects} shows the minimal, the maximal and the average number of objects in a training image for each data set.
\\
For the data set \textit{Inflorescences}, 15 training images were used to learn an accurate CNN model due to the very high number of objects per training image. In average, almost 2000 flower buds can be learned from each training image. An even larger number of training images would cause very long annotation times. Therefore, the number of 15 training images was determined as sufficient enough for this data set.
\\
\newcolumntype{C}[1]{>{\centering\arraybackslash}p{#1}}
\begin{table*}[tb]
  \centering
  \begin{tabular}{p{3.85cm}C{2.5cm}C{1.8cm}C{1.75cm}C{1.75cm}C{1.6cm}}
		\toprule
		Data set & No. of training images &  Max. no. of objects & Min. no.  of objects & Avg. no. of objects \\
		\midrule
		Young Shoots Natural Background & 35 & 20 & 3 & 10.4 \\ \\
		Young Shoots Artificial Background & 34 & 17 & 4 & 10.5 \\ \\
		Inflorescences & 15 & 4451 & 956 & 1956.7 \\ \\
		Pedicels & 40 & 181 & 45 & 94.9 \\ \\
		Grape Berries & 30 & 569 & 66 & 300.4 \\ \\
		Grape Berries Phenoliner & 30 & 378 & 1 & 186.6 \\
    \bottomrule
  \end{tabular}
  \caption{Minimal, maximal and average number of annotated objects in the training images.}
	\label{table-number-objects}
\end{table*}
The data sets \textit{Young Shoots Natural Background} and \textit{Pedi\-cels} show objects which are difficult to learn for the CNN, since the objects and the background have nearly the same colors and since the young shoots have a complicated and varying shape. For these data sets, a higher number of training images is required. Even with the actual number of 35 or 40 training images, the segmentation generated by the CNN is not as accurate as for the other data sets. For these data sets, the number of training images should be set to the maximal limit of an acceptable training and annotation time.
\\
For the remaining data sets (\textit{Young Shoots Artificial Background}, \textit{Grape Berries}, \textit{Grape Berries Phenoliner}), the objects can be well distinguished from the background by the color. But the number of objects per image is not as high as for the data set \textit{Inflorescences} which allows to annotate a higher number of images. For each of these data sets, the number of annotated training images is set to around 30, which was sufficient to achieve accurate results in image segmentation and object localization. In general, the number of training images is a trade-off between the accuracy of the results and the required time to annotate the images.
\\ \\
The task for data sets \textit{Young Shoots Artificial Background} and \textit{Young Shoots Natural Background} is to detect the young shoots. The objective for data set \textit{Inflorescences} is to detect the single flower buds (and so the number of flower buds). Afterwards, the inflorescences can be detected by clustering the single flower buds. For data set \textit{Pedicels}, the objective is to detect the pedicels for calculating their length. For data set \textit{Grape Berries} and \textit{Grape Berries Phenoliner}, the single grape berries should be detected by the system. For data set \textit{Grape Berries Phenoliner}, there are also one-channel infrared images besides the RGB images, so the CNN can be trained on four-channel input images. 
\\ \\
Most importantly, the system works on unprepared images taken by consumer cameras. For this reason, three of the six data sets evaluated in this work (namely, data sets \textit{Young Shoots Natural Background}, \textit{Inflorescences}, and \textit{Grape Berries}) consist of such unprepared field images.
\\
But to show that this general approach also works on images prepared by the user, also three data sets depicting typical applications of prepared images are evaluated, namely, field images with artificial background (here applied to young shoots), images generated by an automated vehicle-based phenotyping platform (here images of ripe berries taken by the Phenoliner platform \cite{kicherer2017phenoliner}), and lab images for specialized phenotyping applications (here applied to pedicels of grapes).\\
For data set \textit{Young Shoots Artificial Background}, white sheets were used to generate a unicolor background for making it possible to detect the young shoots by their color. The Phenoliner was used to generate images with a unicolor background and without natural light influences. As already mentioned, the Phenoliner additionally generates infrared images. Data set \textit{Pedicels} represents images of cropped plant organs taken in a labor.

\subsubsection{Image Annotation}
\label{sec:image_annotation}

\begin{figure}[!t]
  \centering
  \includegraphics[width=0.48\textwidth]{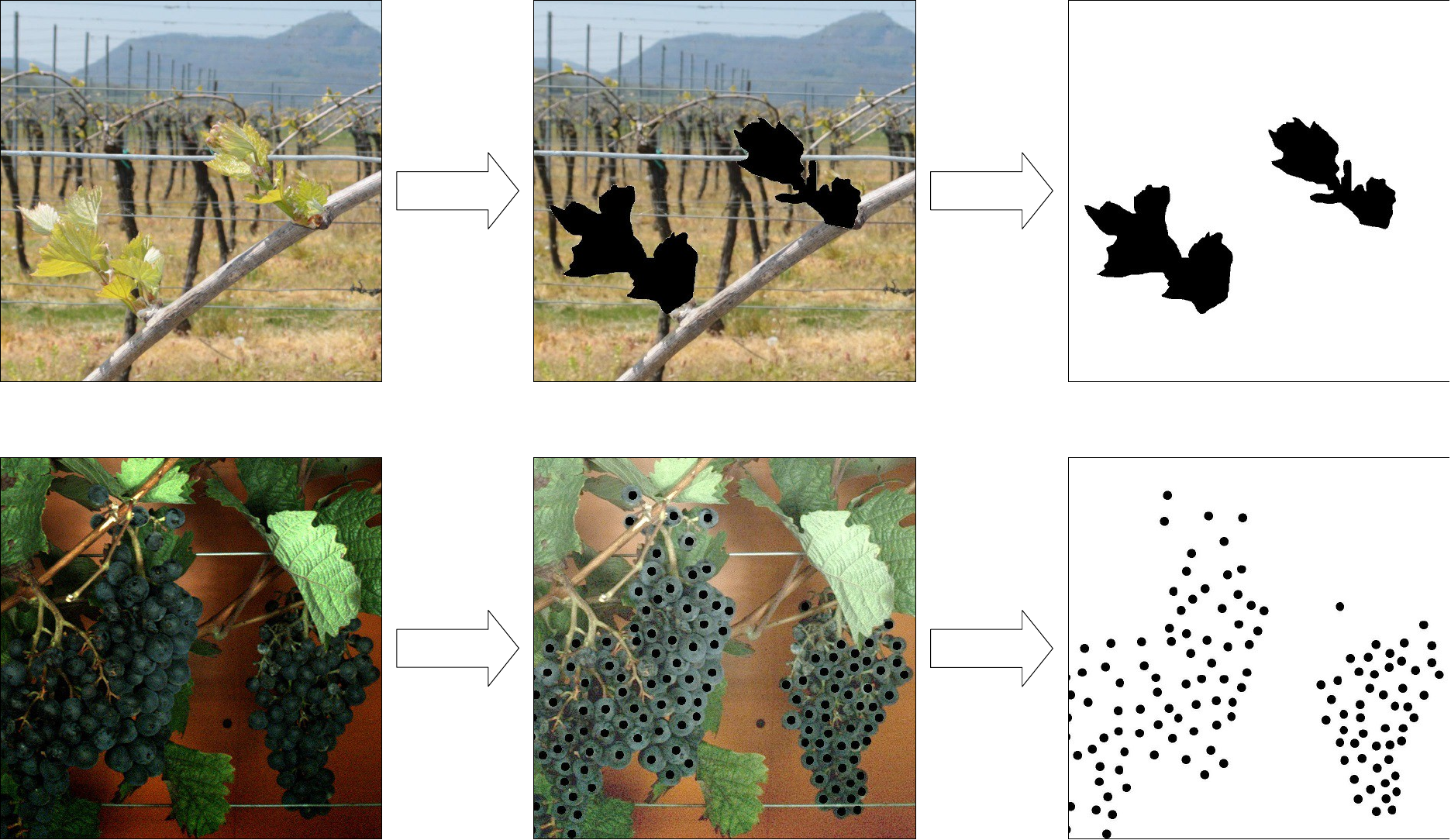}
  \caption{The procedure of annotating images. 
 The target objects can be annotated in black, either with polygons (upper image) or with circles (lower image). Finally, the image is binarized 
  separating black pixels depicting the target objects from brighter pixels depicting non-target object pixels.}
  \label{fig:annotations}
  \vspace*{1ex}
\end{figure}

The output of the trained CNN is a probability map that shows for each pixel the probability to be an object pixel or a background pixel. Since the CNN should learn exactly this task, the ground truth for the training images must be a clear segmentation into object pixels and background pixels. The ground truth of a training image can be represented as a binary image, where object pixels are depicted in black and background pixels in white. To generate such images representing the ground truth, the training images have to be annotated.
\\ \\
The flower buds (data set \textit{Inflorescences}) and the berries (data set \textit{Grape Berries} and \textit{Grape Berries Phenoliner}) are annotated with circles. In order to reduce annotation time, flower buds and berries are annotated with circles having a constant radius for their complete data set. The young shoots are annotated with polygons and the pedicels with lines. 
\\  \\
The annotation process results in a binary image in which all object pixels are black and all background pixels are white (cf. figure \ref{fig:annotations}). 
%


\subsection{Methodology}

\vspace*{1ex}
\subsubsection{CNN-based Segmentation}
\label{sec:cnn_segmentation}

\vspace*{0.5ex}

\paragraph{CNN architecture}
\label{sec:cnn_architecture}

There are already many CNN architectures for semantic segmentation of images. Most of them contain an encoder and a decoder part to generate predictions (output feature maps giving the segmentation) of the original input image size, based on the idea of the Fully Convolutional Networks (FCN) \cite{long2015fully}.
\\
The encoder part is typically a classification network, like AlexNet \cite{krizhevsky2012imagenet} or VGG-Net \cite{simonyan2014very}, but without the final fully-connected layers. Such an encoder contains convolutional layers for learning weights and max pooling layers for reducing the image size.
\\
The decoder contains convolutional layers for learning weights, deconvolutional layers for scaling up the image size and concatenation layers to refine decoder layers with encoder information. This refinement is done by concatenating the feature maps of the decoder layer with feature maps of the encoder layer of the same image size. An alternative to concatenating feature maps is to calculate the pixel-wise sum, as it is done in \citet{long2015fully}. But the advantage of the concatenation layers is that then the network can be trained end-to-end.
\\[1.75ex]
\citet{ronneberger2015u}, \citet{badrinarayanan2017segnet} and \citet{noh2015learning} propose such an encoder-decoder network architecture which can be trained end-to-end. All of them are based on VGG-Net \cite{simonyan2014very} for the encoder part but also differ from each other. DeconvNet \cite{noh2015learning} contains an encoder which is identical to VGG-Net 16 (except the fully connected layers) and a decoder which is symmetric to the encoder. SegNet \cite{badrinarayanan2017segnet} also uses exactly VGG-Net 16 for the encoder part but optimizes the decoder part by reusing pooling indexes from the encoder part for a non-linear upsampling. U-Net \cite{ronneberger2015u} contains a decoder which is symmetric to the encoder, but uses larger image sizes than VGG-Net. Another decoder-based CNN architecture is E-Net \cite{paszke2016enet}, which is optimized for real-time applications by avoiding large VGG-Net components.
\\[1.75ex]
\citet{rudolph2018efficient} use an encoder-decoder network with AlexNet as encoder in order to detect inflorescences in grapevine images. The decoder is similar to that in U-Net, but adapted for a different image size.

\begin{figure*}[bt]
  \centering
  \includegraphics[width=1.0\textwidth]{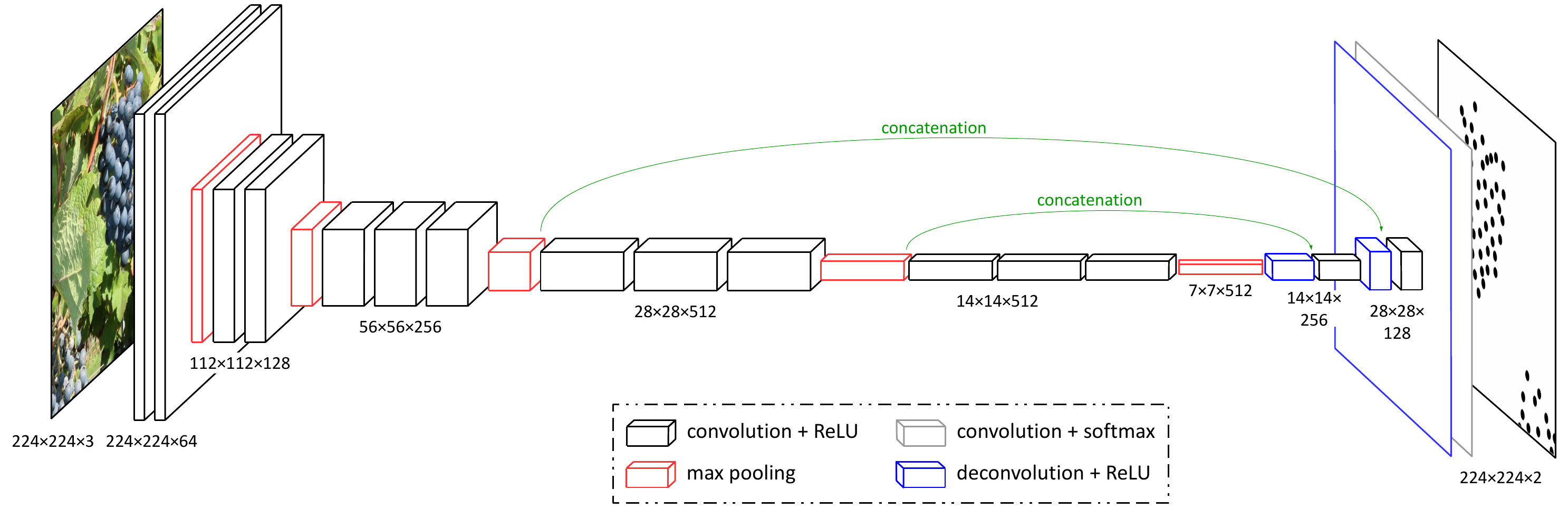}
  \caption{A visualization of the CNN architecture used for this work. }
  \label{fig:cnn}
  \vspace*{2ex}
\end{figure*}

%
The encoder-decoder network for this work uses VGG-Net as encoder, since many related works are also based on VGG-Net and perform well on many popular data sets. An alternative to VGG-Net would be to use AlexNet, which is smaller and faster than VGG-Net. But in terms of IoU score, VGG-Net performs better than AlexNet. This is due to the higher number of learnable weights. For the purpose of this work, a higher performance is more important than a faster running time, since there are some quite complex object types to be learned, as the young shoots in the natural background which has the same colors.
\\[1.75ex]
The decoders of the networks in related works differ more than the encoders. SegNet achieves the best performance on several popular data sets when compared to other networks in literature, therefore the decoder for the CNN used in this work is based on the decoder of SegNet. But there are also two clear differences. SegNet reuses only pooling indexes from the encoder for upsampling in the decoder. U-Net concatenates the whole feature maps of the pooling layer in the encoder to the corresponding feature maps in the decoder upsampled via deconvolution. The variant of U-Net requires more memory but reuses more information of the encoder and therefore promises a better performance. For this reason, the variant of U-Net is used in this work. 
\\[1.75ex]
To reduce the required memory, such a U-Net-based concatenation is only done for the two feature maps with the smallest image size ($7 \times 7 $ and $14 \times 14$). This idea is similar to the decoders of the FCNs, where the information of the corresponding feature map in the encoder is also only reused for the zero (FCN-32), one (FCN-16) or two (FCN-8) smallest feature maps. After these small decoder layers reusing encoder information, a deconvolutional layer with a large stride size (32 for FCN-32, 16 for FCN-16 and 8 for FCN-8) is used to upsample the small feature map to the original image size. FCN-32 performs clearly better than FCN-16, but FCN-16 only slightly better than FCN-8. Since no significant amount of the performance is expected when using additional concatenations of feature maps (similar to FCN-4 or FCN-2), the decoder of the CNN used in this work follows the principle of FCN-8 by reusing encoder feature maps in the decoder for the two smallest feature maps, followed by a deconvolution with 8 strides. In conclusion, the decoder of the CNN used in this work uses ideas and components of FCN-8, U-Net and SegNet.
\\[1.5ex]
E-Net achieves a performance similar to SegNet.
But E-Net is more complex to implement and to adapt due to the higher number of layers, the branched structure and additional components as regularizers with spacial dropout or batch normalization. Additionally, the VGG-based architecture has shown sufficient good run time performance in this phenoyping application (cf.  table \ref{table-test-time} in section \ref{sec:running_time_performance} and is therefore used in this work. 
%
\begin{figure*}[t]
  \centering
  \includegraphics[width=0.8\textwidth]{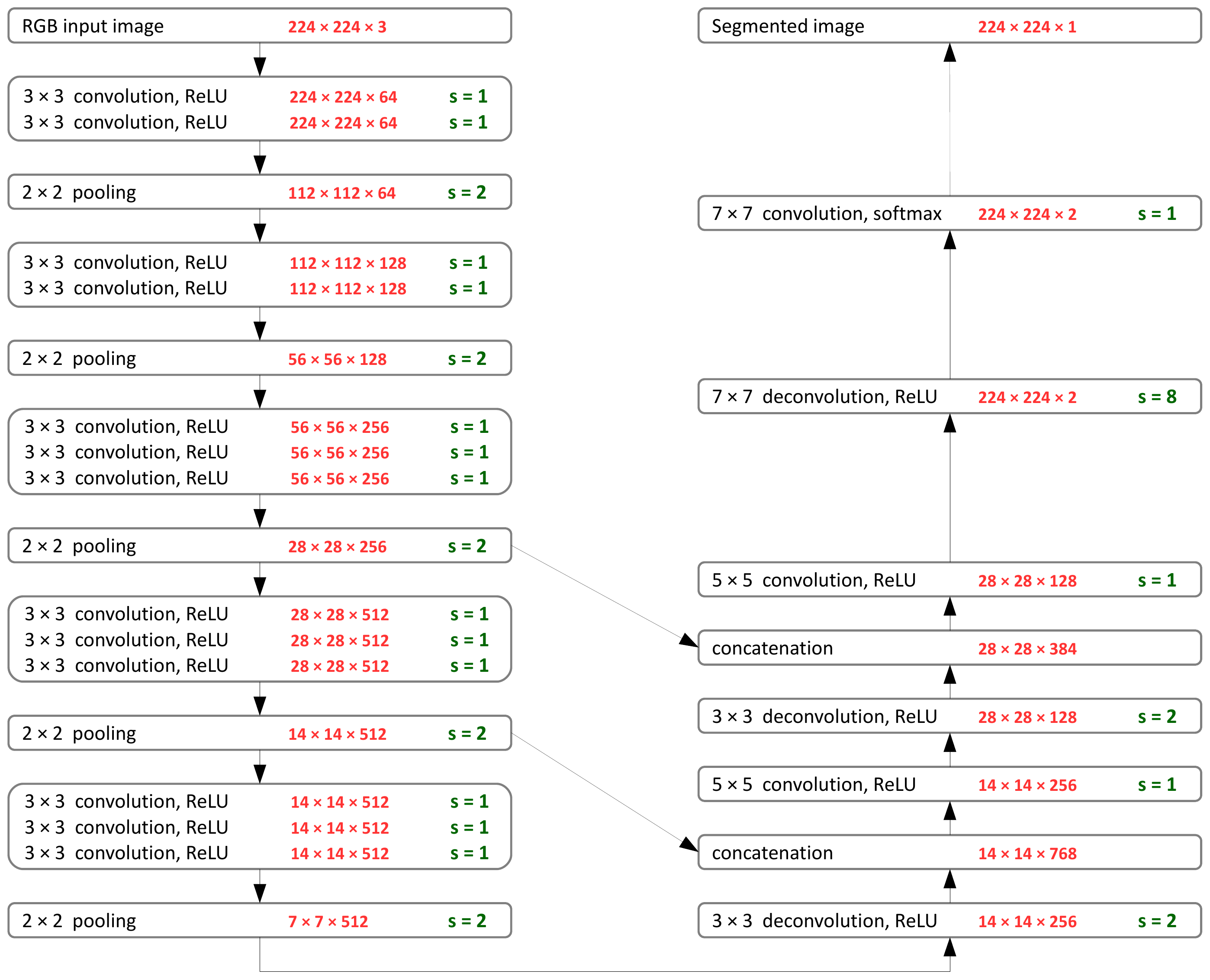}
  \caption{The CNN architecture in more details. The output size of each layer is depicted in red, the number of strides in green. }
  \label{fig:cnn_details}
\end{figure*}
\\[1.25ex]
Now, the CNN architecture used in this work is described in more details. Figure \ref{fig:cnn} visualizes the architecture, figure \ref{fig:cnn_details} gives a more detailed overview about the layers, including kernel size and strides. The encoder part is VGG-Net 16, without the fully-connected layers at the end. The input size of the CNN is $224 \times 224 \times 3$, so it accepts $224 \times 224$ patches of RGB images. For the data set with four input channels (RGB + infrared), the input size is $224 \times 224 \times 4$. In the encoder part, the input images go through five blocks of convolutional layers, each followed by a max pooling layer reducing the image size. For each block, the image size is halved and the number of channels is doubled until reaching a number of 512 channels, starting with 64 channels in the first block. This results in output sizes of $224 \times 224 \times 64$, $112 \times 112 \times 128$, $56 \times 56 \times 256$, $28 \times 28 \times 512$ and $14 \times 14 \times 512$ for the convolutional layers. The final max pooling layer also halves the image size, therefore the output of the encoder has the size $7 \times 7 \times 512$.
\\[1.0ex]
The decoder is related to the FCN-8 described in \citet{long2015fully}, but uses concatenations instead of element-wise summations of layers to enable end-to-end training of the CNN. The output of the encoder goes through a deconvolutional layer doubling the image size ($s=2$) and halving the number of channels, resulting in an output size of $14 \times 14 \times 256$. The output of this deconvolutional layer and the output of the $14 \times 14 \times 512$ max pooling layer are concatenated, hence the input size for the next convolutional layer is $14 \times 14 \times 768$. The subsequent convolutional layer generates an output size of $14 \times 14 \times 256$. The following deconvolutional layer also doubles the image size and halves the number of channels, resulting in an output size of $28 \times 28 \times 128$. Then the output of this deconvolutional layer and the output of the $28 \times 28 \times 256$ max pooling layer are concatenated, thus the input size for the subsequent convolutional layer is $28 \times 28 \times 384$. The output size of this convolutional layer is $28 \times 28 \times 128$. The subsequent deconvolutional layer directly increases the size from $28 \times 28 \times 128$ to the final output size $224 \times 224 \times 2$ by using a stride size of 8 ($s=8$). This step is also done for the FCN-8 and requires a large filter size, like $7 \times 7$ in this architecture. The output of this deconvolutional layer goes through a final convolutional layer with the same output size and the softmax activation function. The final output of the CNN is a 2-channel image having the original image size of $224 \times 224$. The first channel stores for each pixel the probability to be an object pixel (label 0), the second channel stores the probability to be a background pixel (label 1).
\vspace*{-1ex}
\paragraph{CNN Training}
\label{sec:cnn_training}
For training a CNN, training images are needed as well as the corresponding ground truth which should be learned by the CNN. The images representing the ground truth can be created by annotating the training images. The training images have a high resolution of up to $5472 \times 3648$ pixels, but the CNN input size is $224 \times 224$ pixels. Thus the CNN is not fed with whole training images but with image patches of the CNN input size.
\\[1.0ex]
The CNN should be able to generate semantic segmentations of the original image resolution. The high-resolution images have a size of up to $5472 \times 3648$ pixels, the input size of the CNN is only $224 \times 224$ pixels. Adapting the CNN architecture for such a large input size would require much more memory than one could expect in common computers. That is why just image patches are used to train the CNN.
\\[1ex]
There are two possible ways to generate image patches for CNN training. The first way is to sample randomly a fixed number of positions in the image. The patches are generated by cropping image excerpts at these positions. The advantage of this method is, that the user can decide how many patches he wants to generate per image. Its disadvantage is, however, that information is lost, since not the entire image is covered with patches. The second possible way to generate patches is to cover the whole image with adjacent patches. In this case, no information is lost, but the spatial variance is lower since the patch positions are in a static grid and the user can not decide how many patches he wants to generate for each image.\\
%
\begin{figure*}[t]
  \centering
  \includegraphics[width=0.8\textwidth]{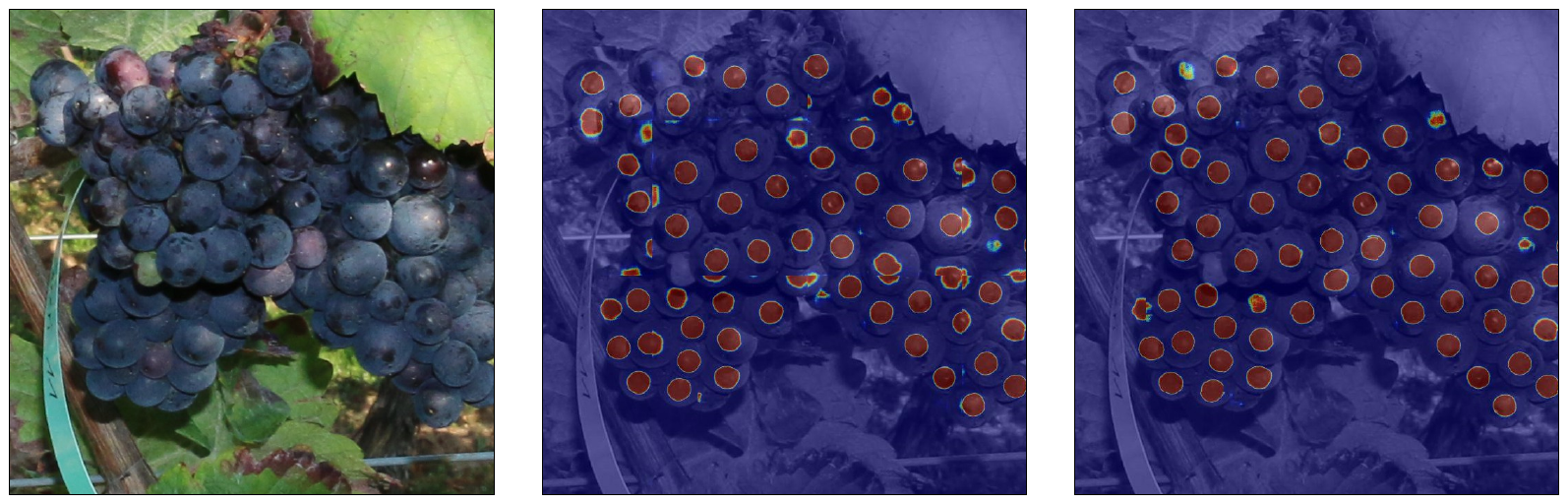}
  \caption{An excerpt of a test image segmented by the CNN. Left: Original image, Middle: Heatmap of the segmentation with non-overlapping patches, Right: Heatmap of the segmentation with patches overlapping by 50\%. Most artifacts at the borders could be removed by using overlapping patches.}
  \label{fig:segmentation_overlap}
\end{figure*}
A common way to increase the number of training images is the so-called data augmentation. Data augmentation means to extend the training image set by modified versions of the image. This makes the learned model more robust against noise and variations. The data augmentation implemented for this work contains three types of modifications: Rotation or flipping, scaling and Gaussian blur. For rotation or flipping, there are six variants: Flipping left-right, flipping top-bottom, transposing the image, rotating by $90^\circ$, rotating by $180^\circ$ and rotating by $270^\circ$. For scaling, there are four different scale factors: $0.8$, $0.9$, $1.1$ and $1.2$. The Gaussian blur is either done with a filter radius of 1 or a filter radius of 2. For each of the three modification types, the variant (or the parameter) is chosen randomly. Since three modified versions of the training image patches are added to the training image set, the data augmentation quadruples the number of training image patches. Experiments on the data set \textit{Grape Berries} show, that each of these three types of data augmentation improves the performance of the CNN in terms of segmenting test images (measured in IoU score).\\ \\
After generating training image patches, the CNN can be trained on these patches using gradient descent. There are three strategies to adjust the weights for the filters while training the CNN: Batch gradient descent, stochastic gradient descent and mini-batch gradient descent.
\\
Batch gradient descent uses the mean error of all training images and updates the weights after each epoch. The cost function for batch gradient descent is smooth and predictable, but the whole data set must be loaded into memory which is not possible for large data sets. Furthermore, it can not be parallelized very well since the weight adjustment depends on all training images.
\\
Stochastic gradient descent updates the weights after each training image after calculating the error by evaluating the cost function. The cost function is noisy, but converges faster than the cost function for batch gradient descent. Stochastic gradient descent is also applicable for large data sets, since only the current training image must be stored in memory, and also a parallelization is possible. But after both cost functions are converged, the cost for batch gradient descent is less then for stochastic gradient descent - that means, that the learned model is more accurate for batch gradient descent.
\\
Mini-batch gradient descent combines the benefits of batch gradient descent and stochastic gradient descent by dividing the training image set randomly into small batches of an equal size, the batch size. Mini-batch gradient descent updates the weights after each batch and uses the mean error of all training images in the current batch. The batch size regulates the trade-off between batch gradient descent (batch size = number of training images) and stochastic gradient descent (batch size = 1). In this work, mini-batch gradient descent is used with a batch size of 20 and a learning rate of $0.0001$. The training cost after 23 epochs varies for the six data sets between $0.001$ and $0.01$, which shows that the cost function is converged well. \\ \\
After the CNN is trained, it can be used to generate a semantic segmentation of unknown test images. The test images must also be divided into patches of CNN input size. After the CNN has segmented each test image patch, the segmented image is constructed by recomposing the patches to an image. The easiest and fastest way to divide a test image into patches is to cover the whole image with adjacent patches. But then, there are artifacts in the segmented image at the borders between the patches (cf. figure \ref{fig:segmentation_overlap}). The number of such artifacts can be reduced by covering the image with overlapping patches, as it is done in \citet{rudolph2018efficient}. But this requires a higher number of patches per image and thereby a higher running time for segmenting a test image. With a patch overlap of 50\% (112 pixels) in both dimensions, the number of patches and thereby the running time is increased by a factor of four. The decision whether it is worth to accept a higher processing time for more accurate results depends on the qualitative difference between a segmentation with overlapping and non-overlapping patches. How large this qualitative difference is depends on the data set.
\subsubsection{Post-Processing and Data Analysis}
\begin{figure*}[!th]
  \centering
  \includegraphics[width=0.8\textwidth]{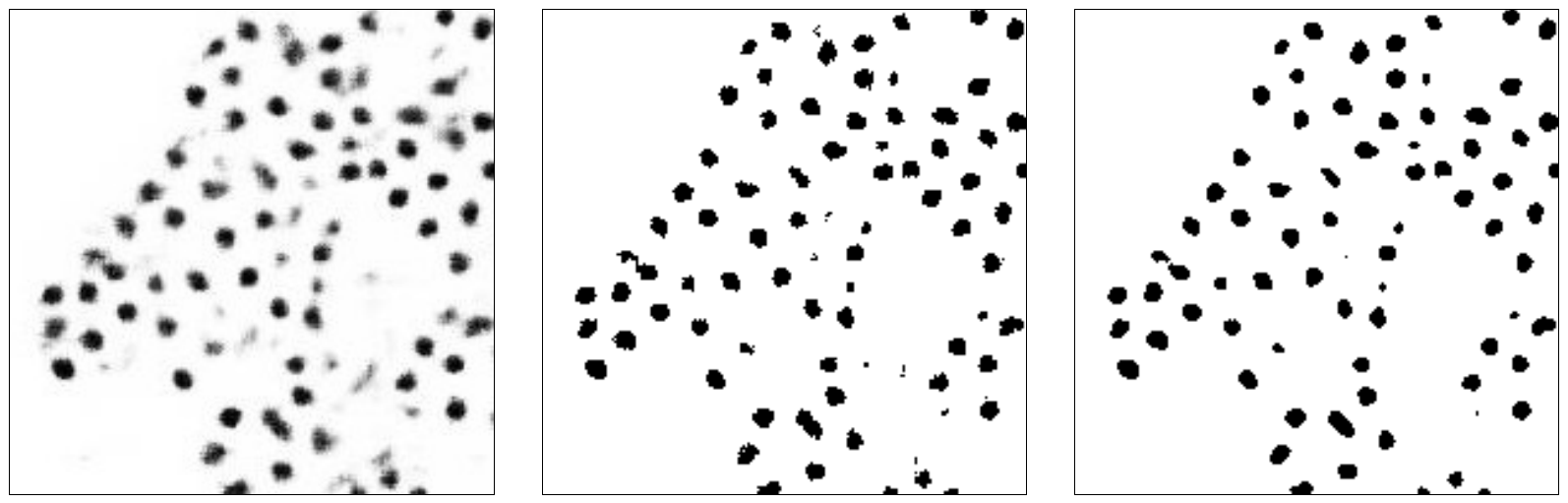}
  \caption{The post processing steps for an image excerpt of the \textit{Inflorescences} data set. The background/object pixel probabilities depicted in grayscales (left), the segmentation after binarization (middle) and after applying the median filter (right).}
  \label{fig:post_processing_steps}
\end{figure*}
\begin{figure*}[!th]
  \centering
  \includegraphics[width=0.8\textwidth]{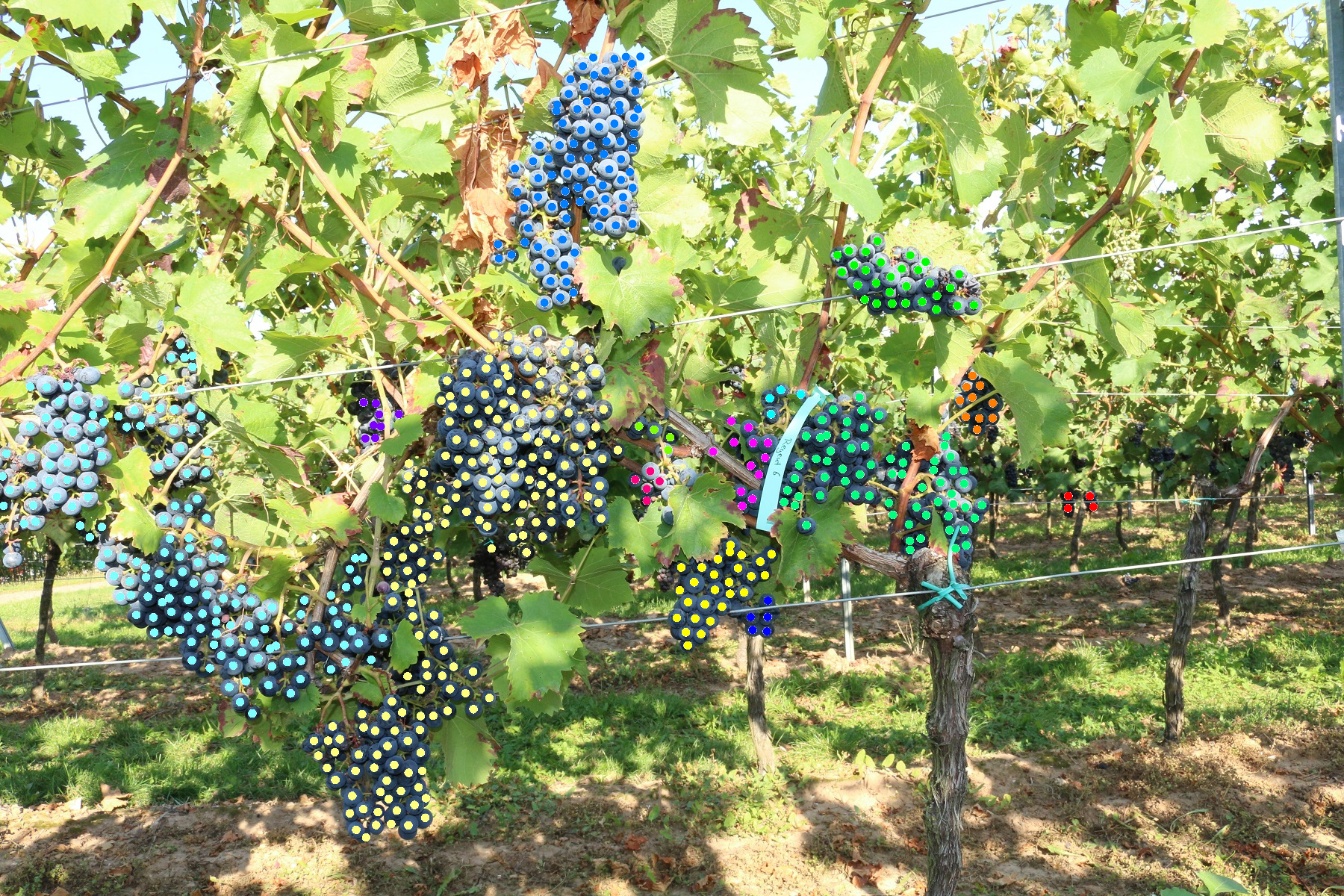}
  \caption{The clustering result for an image of the data set \textit{Grape Berries}. Each color represents a cluster. Most grape bunches are detected correctly as clusters, but there are some errors for overlapping or occluded grape bunches.}
  \label{fig:clusters}
\end{figure*}
\begin{figure}[bth]
  \centering
  \includegraphics[width=0.48\textwidth]{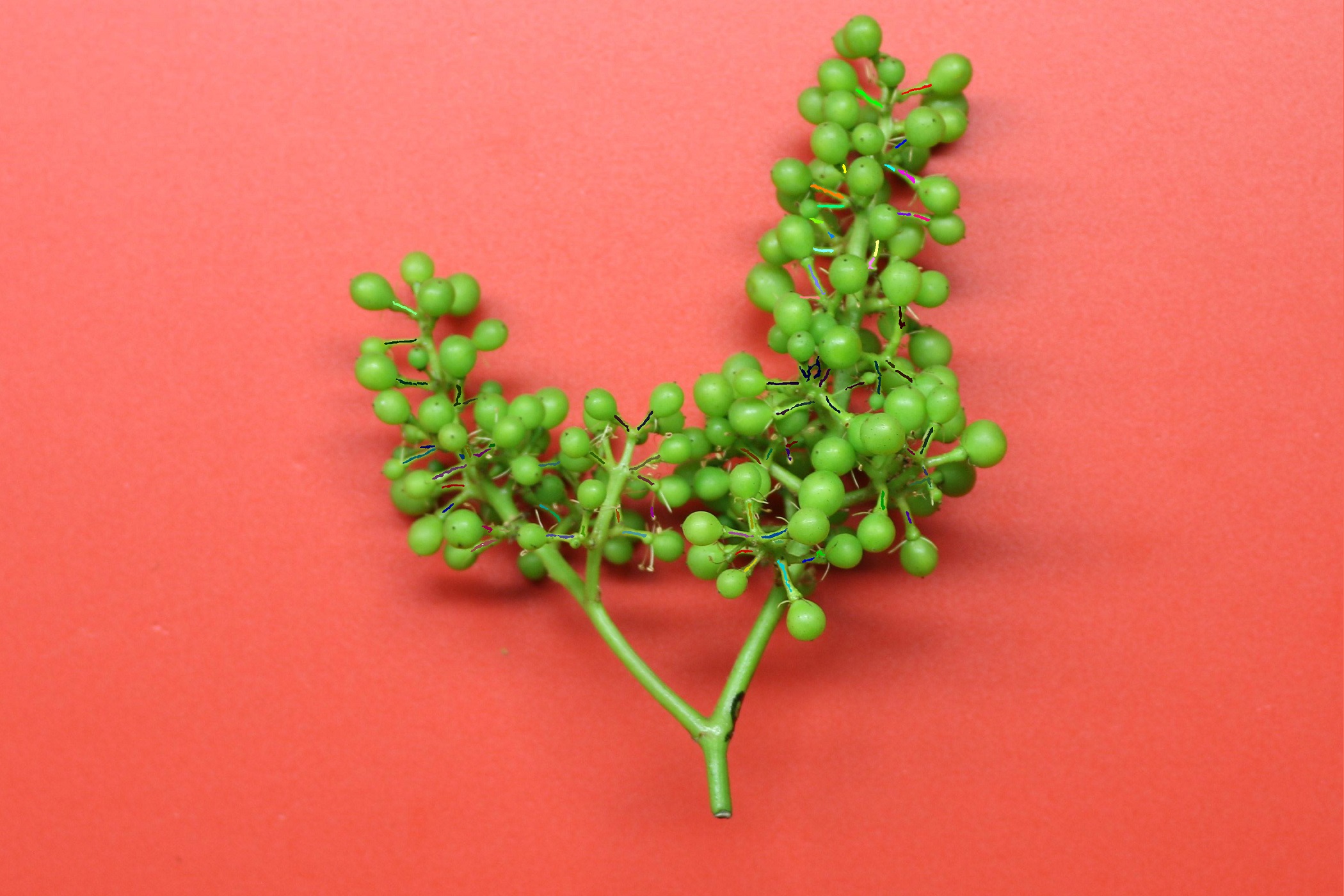}
  \caption{Exemplary image of young grape bunch at BBCH 73 of 'Dornfelder'. Although the F1 score for pedicel detection is only 62.1\%, the most long detected pedicels are true detections. So the results are good enough to calculate the average length of the $n$ most long pedicels.}
  \label{fig:pedicels}
\end{figure}

\label{sec:post_processing}
As already mentioned, the CNN outputs for each pixel the probability to be a background pixel and the probability to be an object pixel, each in one output channel. Since these probabilities are complementary probabilities to each other, only one of them is used for further processing. The CNN output is saved as an image depicting the background pixel probabilities in grayscales between 0 and 255. Figure \ref{fig:post_processing_steps} visualizes the post-processing steps. The first post-processing step is a binarization with a default threshold of 127, resulting in an image in which all object pixels are black and all background pixels are white. The second post-processing step is to apply a $3 \times 3$ median filter on the binarized images to reduce salt-and-pepper noise.
\\
The post-processed semantic segmentation of the test images is used for further data analysis. The most important task for data analysis is to localize and count the individual objects, since the number of objects (like the buds or the grapes) can be used for yield estimation. As already discussed, the objects are annotated as non-overlapping filled circles or polygons, thus the CNN learns that the objects do not overlap. For this reason, the individual objects are also connected components of object pixels in the segmentation generated by the CNN. Connected components can be detected by a region labeling algorithm which assigns an individual segment label to each connected component. Since each connected component represents one object, the detection of connected components enables to determine the number of objects and their positions. Thus, the task of object counting and localization can be done with a simple region labeling algorithm.
\\
Although the median filter reduces noise, some noisy connected components remain after post-processing. Usually, the larger components resemble correctly detected objects and the very small components are noise. For this reason, a parameter \verb|min_region_size| is integrated, that defines a minimal number of pixels for a connected component to be an object. This parameter can be set separately for each data set and regulates the trade-off between false-positive and false-negative detections.
\\ 
For the data sets \textit{Inflorescences}, \textit{Grape Berries} and \textit{Grape Berries Phenoliner}, it is also interesting to compute the number of inflorescences or the number of grape bunches. Furthermore, the number of single flower buds in an inflorescence or the number of berries in a grape bunch can be an important information in viticulture. For this task, a clustering algorithm can be applied on the object centroids of the flower buds or berries. Inflorescences and grape bunches are typically uniform clusters. We assumed that the distance between two adjacent flower buds or berries in the same inflorescence or grape is smaller than the distance between two inflorescences or grapes. For this reason, a distance-based hierarchical clustering was applied. Consequently, a hierarchical clustering algorithm is used with the distance criterion. This means, that a pair of objects corresponds to the same cluster, if and only if the distance between the centroids of the two objects is smaller than an adjustable threshold. In principle, the clustering produces reasonable results on the data sets used in this work. But the standard clustering methods are not robust against overlaps and occlusions (cf. figure \ref{fig:clusters}).
\\
For the data set \textit{Pedicels}, the lengths of the detected pedicels have to be calculated. Since pedicels are straight lines with a certain length, the easiest way to get the length is to determine the extent of the detected pedicels in $x$ and $y$ direction and then calculate the length with the Pythagorean theorem. The system outputs the average length of the $n$ most long pedicels for $n=1$, $n=10$, $n=15$ and $n=N$, where $N$ is the number of detected pedicels. Figure \ref{fig:pedicels} in section \ref{sec:obj_detection_and_localization} visualizes the detected pedicels in a test image. The 15 most long detected pedicels are mostly correct detections, so the calculated lengths are quite accurate for $n=1$, $n=10$ and $n=15$.
%
%
\vspace*{-1ex}
\section{Results and Discussion}
\label{sec:evaluation}
In this section, the evaluation methods and the results of the experiments are presented and discussed. The software is written in Python using the library TensorFlow \cite{tensorflow2015-whitepaper} (version 1.6.0), which contains many functions for implementing neural networks. TensorFlow is optimized for parallelization and GPU support to decrease the running time. TensorFlow requires CUDA Toolkit 9.0 and cuDNN 7.0.5 for the GPU mode, which is used for this work. All computations where done on the following computer:
\begin{itemize}\setlength\itemsep{0pt}
  \item Computer model: Acer Aspire TC-710 (year 2016)
	\item Processor: Intel Core i5-6400 - CPU 2.70 GHz
	\item RAM: 8 GB
	\item Graphic board: NVIDIA GeForce GTX 745 (4 GB DRAM)
	\item Operating system: Windows 10 Home (Build 1803)
\end{itemize}
$\\$
Evaluations were done for the CNN-based image segmentation, for the object detection and localization and for the running time. For each data set, five different evaluations were performed:
\begin{enumerate}
  \item 5-Cover: 5 training images and patches covering the whole training images. Non-overlapping test image patches.
	\item 5-Cover-DA: 5 training images and patches covering the whole training images and data augmentation. Non-overlapping test image patches.
	\item All-Cover: All training images and patches covering the whole training images. Non-overlapping test image patches.
	\item All-Rand-DA: All training images and 100 randomly chosen patches per training image (200 for data set \textit{Inflorescences} since there are only 15 training images) and data augmentation. Non-overlapping test image patches.
	\item Best-Overlap50: That of the four trained CNN models showing the best IoU score (mostly \textit{All-Cover}) with test image patches overlapping by 50\% (112 pixels)
\end{enumerate}
%
%
\subsection{Image Segmentation}
\label{sec:eval_img_segmentation}
Evaluating the image segmentation means to compare the binarized and post-processed segmentation generated by the CNN for the test images with their ground truth. That means, that the test images used for the evaluation must be annotated before, since the ground truth is required for an automated evaluation. The image segmentation is evaluated by calculating the mean intersection-over-union score (mIoU, also known as Jaccard index), since this is the most common evaluation measure for semantic segmentation.
\\ \\
In general, the IoU describes the similarity of two sets $A$ and $B$ and is defined as the quotient of the cardinalities of the intersection and the union of the two sets.
\begin{align}
& \text{IoU}(A, B) = \frac{| A \cap B |}{| A \cup B |}
\end{align}
This definition can be specified for measuring the quality of a semantic segmentation. Let $T_c$ be the set of pixels having the class label $c$ in the ground truth and $P_c$ the set of pixels having the class label $c$ in the segmentation predicted by the CNN. Then the IoU for a class $c$ is defined as follows:
\begin{align}
& \text{IoU}_{c} = \frac{| T_c \cap P_c |}{| T_c \cup P_c |}
\end{align}
The intersection term is equal to the number of pixels of class $c$, which are assigned to the correct class by the CNN, the union term includes also the false positives and false negatives. So the IoU determines which ratio of the pixels of class $c$ are true positives. The mean IoU generalizes the definition by taking the mean over all classes and is a good quality measure for semantic segmentation. The mean IoU is defined as
\begin{align}
& \text{mIoU} = \frac{1}{|C|} \sum_{c \in C} \text{IoU}_{c}
\end{align}
where $C$ is the set of classes. In this work, $C = \{0, 1\}$, where class 0 represents the object pixels and class 1 the background pixels. The definition of $\text{IoU}_{c}$ and mIoU can also be generalized for a whole data set by taking the mean over all segmented images. The mIoU for a data set is often abbreviated with IoU.
\\ \\
\begin{table*}[!t]
  \centering
  \begin{tabular}{llccc}
		\toprule
		Data set & Evaluation & $\text{IoU}_0$ & $\text{IoU}_1$ & mIoU \\
		\midrule
		Young Shoots Natural Background & 5-Cover & 0.0\% & 99.4\% & 49.7\% \\
		& 5-Cover-DA & 36.1\% & 99.6\% & 67.8\% \\
		& All-Cover & 51.3\% & 99.7\% & 75.5\% \\
		& All-Rand-DA & 49.1\% & 99.7\% & 74.4\% \\
		& Best-Overlap50 & 57.8\% & 99.8\% & \textbf{78.8\%} \\
		\midrule
		Young Shoots Artificial Background & 5-Cover & 55.5\% & 99.6\% & 77.5\% \\
		& 5-Cover-DA & 65.7\% & 99.7\% & 82.7\% \\
		& All-Cover & 71.5\% & 99.8\% & 85.6\% \\
		& All-Rand-DA & 66.2\% & 99.7\% & 82.9\% \\
		& Best-Overlap50 & 74.8\% & 99.8\% & \textbf{87.3\%} \\
		\midrule
		Inflorescences & 5-Cover & 35.3\% & 99.7\% & 67.5\% \\
		& 5-Cover-DA & 37.9\% & 99.6\% & 68.8\% \\
		& All-Cover & 42.9\% & 99.7\% & 71.3\% \\
		& All-Rand-DA & 42.3\% & 99.7\% & 71.0\% \\
		& Best-Overlap50 & 44.6\% & 99.8\% & \textbf{72.2\%} \\
		\midrule
		Pedicels & 5-Cover & 13.1\% & 99.8\% & 56.5\% \\
		& 5-Cover-DA & 22.8\% & 99.8\% & 61.3\% \\
		& All-Cover & 28.3\% & 99.8\% & 64.1\% \\
		& All-Rand-DA & 27.9\% & 99.8\% & 63.8\% \\
		& Best-Overlap50 & 30.3\% & 99.8\% & \textbf{65.1\%} \\
		\midrule
		Grape Berries & 5-Cover & 43.9\% & 99.1\% & 71.5\% \\
		& 5-Cover-DA & 50.8\% & 99.2\% & 75.0\% \\
		& All-Cover & 56.1\% & 99.3\% & 77.7\% \\
		& All-Rand-DA & 53.9\% & 99.2\% & 76.5\% \\
		& Best-Overlap50 & 60.4\% & 99.4\% & \textbf{79.9\%} \\
		\midrule
		Grape Berries Phenoliner & 5-Cover & 31.2\% & 98.4\% & 64.8\% \\
		& 5-Cover-DA & 39.1\% & 98.6\% & 68.8\% \\
		& All-Cover & 50.8\% & 98.8\% & 74.8\% \\
		& All-Rand-DA & 53.4\% & 98.9\% & 76.1\% \\
		& Best-Overlap50 & 56.0\% & 99.0\% & \textbf{77.5\%} \\
    \bottomrule
  \end{tabular}
  \caption{Results for the semantic segmentation into object and background pixels measured in IoU score. Object pixels are all pixels corresponding to a plant organ (single flower, young shoot, grape berry or pedicel).}
	\label{table-iou}
\end{table*}
The results for the semantic segmentation are shown in table \ref{table-iou}. As expected, the IoU is higher for the data sets with larger objects (young shoots and grape berries) than for the data sets with smaller objects. The reason for this is that many pixels which are labeled incorrectly by the CNN reside on the borders of the objects and that the number of border pixels is the highest for data sets with small objects. For the same reason, the approach of \citet{rudolph2018efficient} achieves a higher IoU score for the data set \textit{Inflorescences} (87.6\%) than the approach in this work (72.2\%), since \citet{rudolph2018efficient} has annotated coarse polygons around the inflorescences instead of small circles at the single buds.
\\
A comparison of the different evaluations for the same data set shows that the evaluation \textit{5-Cover} shows the lowest IoU scores. The low IoU scores show, that a training set of five images without data augmentation is not big enough to learn complicated models. For the data set \textit{Young Shoots Natural Background}, the $\text{IoU}_0$ score for evaluation \textit{5-Cover} is 0, which means that there are no correct detections of object pixels in the segmentation generated by the CNN. This data set is hard to learn for the CNN, since the young shoots nearly have the same colors as the natural background.\\
The results can be clearly improved by enlarging the small training set of five images using data augmentation (evaluation \textit{5-Cover-DA}). Especially the IoU score of 67.8\% for the data set \textit{Young Shoots Natural Background} indicates, that data augmentation enables to learn segmenting this challenging data set using only five annotated training images.\\
\begin{table*}[!t]
  \centering
  \begin{tabular}{llccc}
		\toprule
		Data set & Evaluation & Precision & Recall & F1 \\
		\midrule
		Young Shoots Natural Background & 5-Cover & 0.0\% & 0.0\% & 0.0\% \\
		& 5-Cover-DA & 29.8\% & 36.8\% & 32.1\% \\
		& All-Cover & 56.7\% & 57.6\% & 55.2\% \\
		& All-Rand-DA & 45.0\% & 55.4\% & 48.6\% \\
		& Best-Overlap50 & 68.9\% & 65.9\% & \textbf{65.6\%} \\
		\midrule
		Young Shoots Artificial Background & 5-Cover & 58.4\% & 67.5\% & 60.1\% \\
		& 5-Cover-DA & 68.7\% & 81.5\% & 73.7\% \\
		& All-Cover & 85.6\% & 77.3\% & 78.7\% \\
		& All-Rand-DA & 58.9\% & 86.4\% & 68.6\% \\
		& Best-Overlap50 & 88.8\% & 77.1\% & \textbf{80.7\%} \\
		\midrule
		Inflorescences & 5-Cover & 80.4\% & 75.6\% & 77.6\% \\
		& 5-Cover-DA & 80.9\% & 81.1\% & 81.0\% \\
		& All-Cover & 92.4\% & 79.9\% & 85.6\% \\
		& All-Rand-DA & 85.7\% & 82.6\% & 84.1\% \\
		& Best-Overlap50 & 93.4\% & 81.1\% & \textbf{86.7\%} \\
		\midrule
		Pedicels & 5-Cover & 61.3\% & 16.6\% & 25.4\% \\
		& 5-Cover-DA & 61.9\% & 42.9\% & 50.1\% \\
		& All-Cover & 69.2\% & 52.8\% & 59.4\% \\
		& All-Rand-DA & 64.4\% & 55.5\% & 59.2\% \\
		& Best-Overlap50 & 70.1\% & 56.3\% & \textbf{62.1\%} \\
		\midrule
		Grape Berries & 5-Cover & 84.6\% & 78.0\% & 80.7\% \\
		& 5-Cover-DA & 81.5\% & 85.1\% & 82.5\% \\
		& All-Cover & 84.5\% & 88.6\% & 85.9\% \\
		& All-Rand-DA & 81.4\% & 88.3\% & 84.1\% \\
		& Best-Overlap50 & 86.6\% & 89.7\% & \textbf{87.6\%} \\
		\midrule
		Grape Berries Phenoliner & 5-Cover & 86.9\% & 58.0\% & 68.5\% \\
		& 5-Cover-DA & 88.8\% & 68.2\% & 76.1\% \\
		& All-Cover & 87.7\% & 83.5\% & 85.4\% \\
		& All-Rand-DA & 89.4\% & 85.2\% & 87.2\% \\
		& Best-Overlap50 & 91.7\% & 85.9\% & \textbf{88.6\%} \\
    \bottomrule
  \end{tabular}
  \caption{Results for the object detection and localization measured in precision, recall and F1 score. The values in the table are the means over all evaluated images of the data set. The detected objects are single plant organs, namely single flower buds, young shoots, grape berries or pedicels.}
	\label{table-obj-localization}
\end{table*}
The evaluation \textit{All-Cover} shows even better results than the evaluation \textit{5-Cover-DA}. This shows, that an enlargement of the data set with additional training images improves the results more than an enlargement with data augmentation. The reason for this is that adding modified versions of the images only helps for generalization but not for collecting more information about the data. Also, the evaluation \textit{All-Cover} mostly shows better results than \textit{All-Rand-DA} for the same reason: Information is lost by taking random image patches instead of covering the whole image with patches. The only exception is data set \textit{Grape Berries Phenoliner}, whose images have a lower resolution. For this resolution, 100 random patches nearly cover the whole image, so the amount of the IoU score can be explained by the data augmentation.\\
On the other hand, this part of the evaluation is only taking into account the IoU scores.  The evaluation of the running time performances in section \ref{sec:running_time_performance} will show that better IoU scores using \textit{All-Cover} must be paid by higher amounts of training time.
Evaluation \textit{Best-Overlap50} shows the best results for all data sets, since the segmentation is more accurate without the artifacts at the borders between the patches.
\subsection{Object Detection and Localization}
\label{sec:obj_detection_and_localization}
The performance of the approach for object detection and localization can be measured by the standard measures Precision, Recall and F1 score. 
%
For the task of object localization, the definition of true and false detections is not trivial. True positives are objects in the ground truth, which are also detected by the CNN at nearly the same position. False positives are objects detected by the CNN, for which no object at nearly the same position exists in the ground truth. False negatives are objects in the ground truth, which are not detected by the CNN at nearly the same position. Hence, the definition of true and false detections depends on the definition of ``nearly the same position''. This can be defined as follows: An object in the ground truth and an object in the segmentation generated by the CNN have nearly the same position, if and only if the euclidean distance between their centroids is smaller than $t$ pixels. Since each object is a connected component of object pixels, the centroid can be calculated easily. The parameter $t$ is called \verb|obj_cnt_tolerance| in the source code and in the documentation. This parameter should be chosen dependent on the object diameter $d$. For data sets for which the objects are annotated as circles of the same radius $r$, the diameter can be easily calculated as $d=2r+1$. For data sets for which the objects are annotated as polygons with different sizes, the diameter can be set to the global average object diameter for this data set or can be estimated dependent on representative objects. 
%
\\ 
%
\begin{table*}[!t]
  \centering
  \begin{tabular}{p{3.0cm}C{2.5cm}C{2.0cm}C{2.5cm}C{2.5cm}C{2.5cm}}
		\toprule
		Data set & Image resolution (px) & No. of Training images & Training time 5-Cov-DA & Training time All-Cov \\
		\midrule
		Young Shoots Natural Background & $5472 \times 3648$ & 35 & 27min 49s & 9h 17min & 16h 12min \\ \\
		Young Shoots Artificial Background & $3456 \times 2304$ & 34 & 10min 40s & 3h 34min & 6h 01min \\ \\
		Inflorescences & $5472 \times 3648$ & 15 & 27min 15s & 9h 03min & 6h 50min \\ \\
		Pedicels & $5472 \times 3648$ & 40 & 27min 32s & 9h 10min & 18h 23min \\ \\
		Grape Berries & $5472 \times 3648$ & 30 & 26min 55s & 8h 58min & 13h 28min \\ \\
		Grape Berries Phenoliner & $2592 \times 2048$ & 30 & 7min 13s & 2h 25min & 3h 35min \\
    \bottomrule
  \end{tabular}
  \caption{CNN training times for the six evaluated data sets. The training time depends on the number of training images and on the image resolution. The training time for 5-Cov-DA is about 20 times the training time for one training image (5 training images and $\times$ 4 data augmentation), the training time for All-Cov (all training images) is about the training time for one training image multiplied by the number of training images. }
	\label{table-training-time}
\end{table*}
Table \ref{table-obj-localization} depicts the results of object detection and localization in terms of precision, recall and F1 score.
As expected, the best results (F1 score) were achieved for the data sets \textit{Grape Berries Phenoliner}, \textit{Grape Berries}, \textit{Inflorescences} and \textit{Young Shoots Artificial Background}. The grapes have a simple shape and a color which can be distinguished well from the background, the young shoots with artificial background can be directly detected by the color since the artificial background is white. The buds in the inflorescences have a color similar to the background, but there is a high number of buds in each image (mostly between 1000 and 3000). The high number of learned examples explains the good results (F1 = 86.7\%) for flower bud detection and localization. The results are clearly better than in the work of \citet{rudolph2018efficient}, where an F1 score of 75.2\% was reached. This difference shows, that the CNN performs better in the task of flower bud detection in images of inflorescences than the circular Hough transform in regions of interest detected by the CNN.
The observations obtained by comparing the different evaluations for the same data set are the same as for the evaluation of the segmentation: The larger the number of training images, the better the results for object detection and localization. Data Augmentation also helps to improve the results, but not as much as using additional training images, since new images contain more information about the objects than modified versions of old images. The usage of overlapping test image patches improves the results due to the more accurate segmentation generated by the CNN.

\subsection{Run Time Performance}
\label{sec:running_time_performance}
%
\begin{table*}[!t]
  \centering
  \begin{tabular}{p{3.0cm}C{3.5cm}C{3.5cm}C{3.5cm}C{3.5cm}}
		\toprule
		Data set & Annotation time (per image) & No. of images to annotate & Total annotation time \\
		\midrule
		Young Shoots Natural Background & 8 min & 59 (35+24) & 7h 52min  \\ \\
		Young Shoots Artificial Background & 7 min & 68 (34+34) & 7h 56min  \\ \\
		Inflorescences & 35 min & 30 (15+15) & 17h 30min \\ \\
		Pedicels & 4 min & 80 (40+40) & 4h 0min \\ \\
		Grape Berries & 9 min & 60 (30+30) & 9h 0min \\ \\
		Grape Berries Phenoliner & 7 min & 60 (30+30) & 7h 0min \\
    \bottomrule
  \end{tabular}
  \caption{Required time for annotating an image for each data set}
	\label{table-annotation-time}
\end{table*}
\begin{table}[tbh]
  \centering
  \begin{tabular}{p{3cm}C{2.5cm}C{2.5cm}}
		\toprule
		Data set & Segmentation time (non-overlapping patches) & Segmentation time (overlapping patches)\\
		\midrule
		Young Shoots Natural Background & 2 min & 8 min \\ \\
		Young Shoots Artificial Background & 45 s & 3 min \\ \\
		Inflorescences & 2 min & 8 min \\ \\
		Pedicels & 2 min & 8 min \\ \\
		Grape Berries & 2 min & 8 min \\ \\
		Grape Berries Phenoliner & 30 s & 2 min \\
    \bottomrule
  \end{tabular}
  \caption{Times that the CNN needs to segment a test image}
	\label{table-test-time}
\end{table}
The run time for training the CNN depends on the image resolution and the number of training image patches. Table \ref{table-training-time} compares the training time for a small set of five training images and a larger set of all training images of the data set, both with patches covering the whole training images. Data augmentation is only done for the sets of five training images, since it quadruples both the number of patches and thus the training time. The maximal training time of about 18 hours is needed for the data set \textit{Pedicels}, for which the larger training set consists of 40 images of a high resolution. The number of training images is limited due to the training time, but only if the number of training images is very high. Even for 100 training images of the highest resolution ($5472 \times 3648$ px), the training time would be about two days and thus still practicable for usual consumer computers as used for this work.
\\
Moreover, the number of training images is also limited due to the annotation times. Table \ref{table-annotation-time} shows the annotation times per image for each data set. Considering a large set of 100 training images, the total annotation time for the data set \textit{Inflorescences} (35 minutes per annotated image) would be about 58 hours, which is very much if only one person annotates the images. But also for the data set \textit{Grape Berries Phenoliner} (7 minutes per annotated image), the total annotation time would be about 12 hours, which requires almost two working days for annotations. But since the results are already good for smaller training sets of maximal 40 images, the training as well as the annotations are practicable in an acceptable time.

Compared with the work of \citet{rudolph2018efficient}, the annotation times are long in this work. \citet{rudolph2018efficient} annotate coarse polygons around the inflorescences, which are regions of interest for flower bud detection. This requires only an annotation time of around one minute per image, which is very little compared with the time of 35 minutes needed for annotating each single bud. But the benefit of the accurate annotations of each single object is, that the CNN can be directly trained end-to-end on object detection and localization. The results show, that the more accurate approach in this work achieves a clearly higher performance in flower bud detection in field images of inflorescences than the less accurate approach of \citet{rudolph2018efficient}. The F1 score of 86.7\% achieved in this work is clearly higher than the F1 score of 75.2\% achieved in the work of \citet{rudolph2018efficient} for flower bud detection in ROIs predicted by the CNN, and is also higher than the F1 score for flower bud detection in the ground truth of the ROIs (80.0\%). In summary, the approach used in this work requires more time for image annotation than the approach of \citet{rudolph2018efficient}, but it is more accurate and therefore performs better. 
\\
The time for segmenting one test image is shown in table \ref{table-test-time}. The time for post-processing and data analysis ranges between 5 and 10 seconds per image and therefore can be neglected. Dependent on the number of test images, the image resolution and the computational capacities, it can be decided to use the faster and slightly less accurate variant with non-overlapping test images patches or to use the slower but slightly more accurate variant with overlapping test image patches. 


\section{Conclusions}
The objective of this work was twofold. First, we aimed for the development of a general approach to enable automated detection, localization and counting of plant organs of grapevines in different growth stages to avoid development and programming of specialized software for each growth stage. Second, we aimed for an approach that is able to perform robust detection, localization and counting in unprepared field images taken by consumer cameras. \\
To meet both challenges, we employ a CNN-based semantic segmentation that is end-to-end learnable and therefore allows learning visual features of plant organs given different growth stages and backgrounds directly by global optimization. The performance of the approach was evaluated for the CNN-based semantic segmentation (IoU score) and for the object detection and localization (F1 score). The approach was evaluated on six different data sets. Three of them contain unprepared field images taken by consumer cameras. Three additional data sets are evaluated to demonstrate that the approach can also successfully applied to prepared images taken (1) in the field with artificial background, (2) in the field by images generated from a mobile phenotyping platform, and (3) in the lab for specialized phenotyping purposes. For the segmentation, IoU scores of 65.1\% up to 87.3\% were achieved. For the object detection and localization F1 scores of 62.1\% up to 88.6\% were achieved. \\
All in all, data acquisition hereby is simple-to-apply enabling high-throughput, non-invasive field phenotyping (except for the study case of pedicels) of large breeding populations, genetic repositories or mapping populations where received objective and precise phenotypic data. The determination of the length of pedicels is challenging also under lab conditions due to the complex grape bunch architecture and overlapping objects. The workflow works on standardized images of individual grapes in front of an artificial background. The lengths of pedicels are correlated to looser grape bunch architecture which is one of the most important selection criteria in breeding programs. Non-invasive image capture of individual grape bunches directly in the field (with natural or artificial background) will also enable monitoring experiments within large mapping populations for genetic studies and genetic marker development due to Quantitative Trait Loci (QTL) analysis.

\bibliography{references}

\end{document}